%% file: main.tex
\documentclass{article}
\usepackage{microtype}
\usepackage{graphicx}
\usepackage{subfigure}
\usepackage{amsmath}
\usepackage{stmaryrd}
\usepackage{booktabs} 
\usepackage{amssymb}
\usepackage{hyperref}
\usepackage{soul}
\usepackage{xcolor}

\newcommand{\cameraready}[2]{#2}

\usepackage{mathtools}
\usepackage{shortcuts}
\usepackage{amsmath}
\usepackage{sidecap}

\usepackage[accepted]{icml2021}

\icmltitlerunning{Disentangling Syntax and Semantics in the Brain with Deep Networks}

\begin{document}
\twocolumn[

\icmltitle{Disentangling Syntax and Semantics in the Brain with Deep Networks}

\icmlsetsymbol{equal}{*}

\begin{icmlauthorlist}
\icmlauthor{Charlotte Caucheteux}{to,goo}
\icmlauthor{Alexandre Gramfort}{to}
\icmlauthor{Jean-Remi King}{goo,ed}
\end{icmlauthorlist}

\icmlaffiliation{to}{Inria, Saclay, France}
\icmlaffiliation{goo}{Facebook AI Research, Paris, France}
\icmlaffiliation{ed}{École normale supérieure, PSL University, CNRS, Paris, France}
\icmlcorrespondingauthor{Charlotte Caucheteux}{ccaucheteux@fb.com}
\icmlkeywords{Neuroscience, Natural Language Processing, Deep Learning, Graph Embedding, Explainable AI}

\vskip 0.3in
]

\printAffiliationsAndNotice{}

\begin{abstract}
The activations of language transformers like GPT-2 have been shown to linearly map onto brain activity during speech comprehension. However, the nature of these activations remains largely unknown and presumably conflate distinct linguistic classes. 
Here, we propose a taxonomy to factorize the high-dimensional activations of language models into four combinatorial classes: lexical, compositional, syntactic, and semantic representations. We then introduce a statistical method to decompose, through the lens of GPT-2's activations, the brain activity of 345 subjects recorded with functional magnetic resonance imaging (fMRI) during the listening of \texttildelow 4.6 hours of narrated text. 
The results highlight two findings. First, compositional representations recruit a more widespread cortical network than lexical ones, and encompass the bilateral temporal, parietal and prefrontal cortices. Second, contrary to previous claims, syntax and semantics are not associated with separated modules, but, instead, appear to share a common and distributed neural substrate. Overall, this study introduces a \cameraready{general}{versatile} framework to isolate\cameraready{}{, in the brain activity,} the distributed representations of linguistic constructs\cameraready{ generated in naturalistic settings}.
\end{abstract}

\section{Introduction}

\input{figures}

Within less than three years, transformers have enabled remarkable progress in natural language processing \cite{devlin_bert_2019,radford_language_nodate}. Pretraining these architectures on millions of texts to predict words from their context greatly facilitates translation, text synthesis and the retrieval of world-knowledge \cite{lample_cross-lingual_2019, brown_language_2020}.

Interestingly, the activations of language transformers tend to linearly map onto those of the human brain, when presented with the same sentences \cite{jain2018incorporating, toneva2019interpreting, abnar_blackbox_2019, schrimpf_artificial_2020, caucheteux2020language, goldstein2021thinking}. This linear mapping suggests that, in spite of their vast learning\footnote{The brain learns continuously from a small set of situated sentences, whereas transformers learn from large sets of pure texts.} 
and architectural differences\footnote{The brain is a single-stream recurrent architecture, whereas the transformer is a multi-stream feedforward architecture.}, 
the brain and language transformers converge to similar linguistic representations \citep{caucheteux2020language,caucheteux2021gpt}. 

However, the nature of these shared representations \cameraready{is}{remains} largely unknown. Three factors explain this gap-of-knowledge.
First, linguistic theories are generally described and interpreted in terms of combinatorial \emph{symbols} (discrete words, syntactic trees, etc). In contrast, brain and language transformers generate high-dimensional \emph{vectors} (a.k.a ``distributed" representations). While these formats are formally equivalent \cite{smolensky_tensor_1990}, interpreting vectorial representations in language models and in the brain \cameraready{remains}{is particularly} challenging.

Second, the representations of deep learning models have been interpreted independently of brain imaging. For example, deep neural networks have been shown to encode lexical analogies in their word embeddings \cite{mikolov_distributed_2013}, as well as singular/plural relationships \cite{lakretz_emergence_2019}, long-distance dependency information \cite{jawahar_what_2019}, and syntactic trees \cameraready{in their middle layers }{}\cite{manning_emergent_2020}. 
    Similarly, the \cameraready{human }{}brain responses to language have been decomposed into a cascade of representations, which maps speech and reading input into phonetic (or orthographic), morphemic, lexical, and syntactic representations \cite{hickok_cortical_2007,  dehaene_unique_2011, pallier2011cortical, friederici_brain_2011, mesgarani2014phonetic, huth_natural_2016, nelson_neurophysiological_2017,brennan2019hierarchical, gwilliams2020neural}. 
  However, we do not know whether all or any of these representations effectively drive the linear mapping between language models and the brain.


Third, the mapping between language transformers and the brain has been mainly investigated with speech and/or narratives \cite{schrimpf_artificial_2020,toneva2019interpreting,abnar_blackbox_2019,reddy2020syntactic} (although see \cite{caucheteux2020language}). \cameraready{These uncontrolled conditions}{The resulting sentences are thus poorly controlled and} potentially confound various features such as phonological variations, sentiment contours, semantic contents, and syntactic properties (\emph{e.g.} stressful texts may tend to be read more quickly, and make use of smaller constituency trees). In sum, the linear correspondence observed between language models and the brain may be driven by a \cameraready{}{wide} variety of factors.

Here, we aim to decompose the similarity between the brain and high-performance language transformers like GPT-2 \cite{radford_language_nodate}, in light of four distinct linguistic classes, namely lexical, compositional, syntactic and semantic representations. To that end, we 
\cameraready{propose an operational taxonomy to separate them in vector systems.}{formalize a taxonomy that factorizes them into four distinct vector bases}. 
We then describe a statistical procedure to \cameraready{estimate distributed syntactic representations}{extract syntactic representations from neural networks, decompose their lexical and compositional components, and separate them from semantic representations}.
Finally, we assess the linear mapping between i) the factorized activations of GPT-2 and ii) the brain signals of 345 subjects listening to the same narratives (4.6 hours of audio stimulus in total) as recorded with functional magnetic resonance imaging (fMRI) \citep{nastase_narratives_2020}.

\section{Operational Taxonomy}
\label{taxonomy}

The notions of lexicon, composition, syntax and semantics are notoriously debated in linguistics. Without pretending to \cameraready{solve this issue}{resolve these debates}, we propose 
\cameraready{\emph{operational} definitions designed to decompose these four classes within vectorial systems.}{five definitions that unambiguously decompose the distributed representations of artificial and biological neural networks.}

First, we use the standard definition of a \emph{representation} as the \cameraready{linearly-extractible} information \cameraready{}{that can be linearly extracted from} a vector of activations, with the rationale that a single artificial or biological neuron can read-out this information \cite{kriegeskorte_representational_2008,king2018encoding}.
In this view, a system $\Psi_1$ is said to share the representation of a system $\Psi_2$ if there exists a linear mapping from $X$ to $Y$, where $X=\Psi_1(w)$ and $Y=\Psi_2(w)$ are the activations elicited by the words $w$ in each system.

Second, we define \emph{lexical} representations as the representations that are context-invariant. This definition follows the standard notion of (non-contextualized) word-embeddings, which associate a unique vector to each word of a dictionary. 

By contrast, we define  \emph{compositional} representations as the ``contextualized'' representations generated by a system combining multiples words: $\Psi(w_1 \dots w_M)$. For clarity, we restrict the term ``compositional" to its strict sense: \emph{i.e.} to the set of representations that cannot be accounted for by lexical representations, and thus by a linear combination of word-embeddings. 


Fourth, we define \emph{syntactic} representations as the set of representations associated with the structure of sentences \cameraready{}{independently of their meaning}. 
Linguistic theories have proposed symbolic representations of such structures (e.g part-of-speech, dependency and constituency trees, see Figure \ref{fig:fig1}). 
\cameraready{Here, we introduce a vectorial representation of syntax through the length of deep neural transformers, following the idea that transformers linearly encode syntactic properties \cite{jawahar_what_2019} such as syntactic distances \cite{manning_emergent_2020}}{Furthermore, deep language models have been shown to linearly encode some of these features  \cite{jawahar_what_2019,manning_emergent_2020,lakretz_emergence_2019,lakretz2020limits,linzen2020syntactic}. Here, we introduce a versatile method to extract the distributed representations of syntax in a deep language model}.
\cameraready{We will build those}{Specifically, we extract these syntactic} representations \cameraready{out of}{from} the average \cameraready{of the }{}activations elicited by a set of synthetic sentences that share the same syntactic properties \cameraready{}{ (Section~\ref{embed_synax}}).

Finally, even though a variety of meaningful features \cameraready{appeared to be}{are} captured by both word embeddings \cite{mikolov_distributed_2013} and contexualized embeddings \cite{radford_language_nodate}, meaning and semantics  \cameraready{remain}{are notoriously} difficult to define formally \cite{jackendoff_foundations_nodate}. To decompose syntax and semantics in distributed representations, we thus propose to define \emph{semantic} representations as the \cameraready{}{lexical or supra-lexical} representations of a language system that are not syntactic.

According to these \cameraready{}{five} definitions, lexical and compositional classes \cameraready{thus} fully decompose both syntax and semantics (and \emph{vice versa}). For example, lexico-syntactic representations refer to the functional categories of words \cameraready{: the }{(}part-of-speech \cameraready{(}{}\emph{i.e.} verb, noun, adjective, \emph{etc.}). 
By contrast, compositional syntax refers to the representations that link words with one another, typically referred to as dependency (or constituency) trees. \cameraready{Similarly}{For example}, in the phrase \textsc{not very happy} (Figure \ref{fig:fig1}), the \cameraready{}{set of} lexical meaning\cameraready{ of words (\emph{i.e.} the linear combination of each of the three words)}{} can be distinguished from their compositional meaning\cameraready{ (\emph{i.e.} $\approx$ sad)}. 
\cameraready{This meaning of the composition need not represent syntax}{The representation of this composition need not contain syntactic information}, \cameraready{in that it could have been (approximately) generated with another sentence}{because its outcome ($\approx$\textsc{sad}) can be similar across phrases following distinct syntactic structures} (\emph{e.g.} \textsc{not very happy $=$ down in the dumps $=$ somewhat sad}, \emph{etc.}). 
\cameraready{Critically}{Note that}, under this definition, \cameraready{both transformers and the brain may generate}{the} distributed representations of syntax \cameraready{that have not been theorized }{need not have a symbolic counterpart in theoretical linguistics} \cameraready{yet}{} -- \emph{e.g.} temporary structures that allow building the syntactic tree of a sentence\cameraready{}{, represent multiple alternative and their respective probabilities \emph{etc}}.




\section{Methods}
\subsection{Isolating Syntactic Representations}
\label{embed_synax}
\input{figures2}
\cameraready{Here, we introduce}{We introduce below} a method to isolate distributed representations of syntax \cameraready{}{in neural networks}. We assume that a system $\Psi$ ($\Psi : \mathcal{V}^M \rightarrow \mathbb{R}^{d\times M}$, $\mathcal{V}$ a vocabulary of words), takes sequences of $M$ words as inputs and generates activations that encode syntactic properties (among other properties). 

Let $w$ be a sentence of $M$ words ($w \in \mathcal{V}^M$, e.g \textsc{the cat is on the mat}), and $\Omega_w$ be the set of sentences that have the same syntax as $w$ \cameraready{}{(\emph{e.g.} \textsc{a boy goes to a pool}, \textsc{this boat floats near the shore}, \emph{etc.})}. The syntactic representation of $w$ is, by construction, also the syntactic representations of all sentences $w' \in \Omega_w$.  If this common syntactic representation is denoted $\overline{\psi} \in \mathbb{R}^d$ , we have:
\begin{align*}
\forall w' \in \Omega_w, \quad \Psi(w')  =\overline{\psi} +  z_{w '}
\end{align*}
with $z_{w'}$ a random perturbation of distribution $\mathbb{P}_w$, that corresponds to the non-syntactic part of the randomized activations $\Psi(w')$. If the density of $\mathbb{P}_w$ is well-defined and centered around 0, then: 
\begin{align*}
\mathbb{E} \big[ \Psi(w') \big] & =  \overline{\psi} \enspace ,
\end{align*}
where $w'$ is sampled uniformly in $\Omega_w$.
Thus, $\overline{\psi}$ (the syntactic representation of $w$) can be approximated through:
\begin{align*}
\overline{\Psi}_k  & = \frac{1}{k} \sum_{i=1}^{k}\big( \overline{\psi} +  z_{w_i}\big)
\xrightarrow[k \to \infty]{l.l.n} \overline{\psi}
\end{align*}
with $(z_{w_1}, \dots, z_{w_k})$ $i.i.d$ samples from $\mathbb{P}_w$. 

\cameraready{}{Overall, the syntactic component of the activations is the average of activations induced by random sentences of the same syntax (Figure \ref{fig:fig2}).}

\subsection{Mapping Representations onto FMRI Signals}
\label{mapping}

\input{mapping_new}
\label{sec:mapping}

\subsection{Decomposing Shared Activations between Brains and Neural Language Models}
\label{decomposing}

\input{decomposing_new}
\input{methods}

\label{expe}

\begin{figure}[ht!]
\vskip 0.1in
\centering
\includegraphics[width=\linewidth]{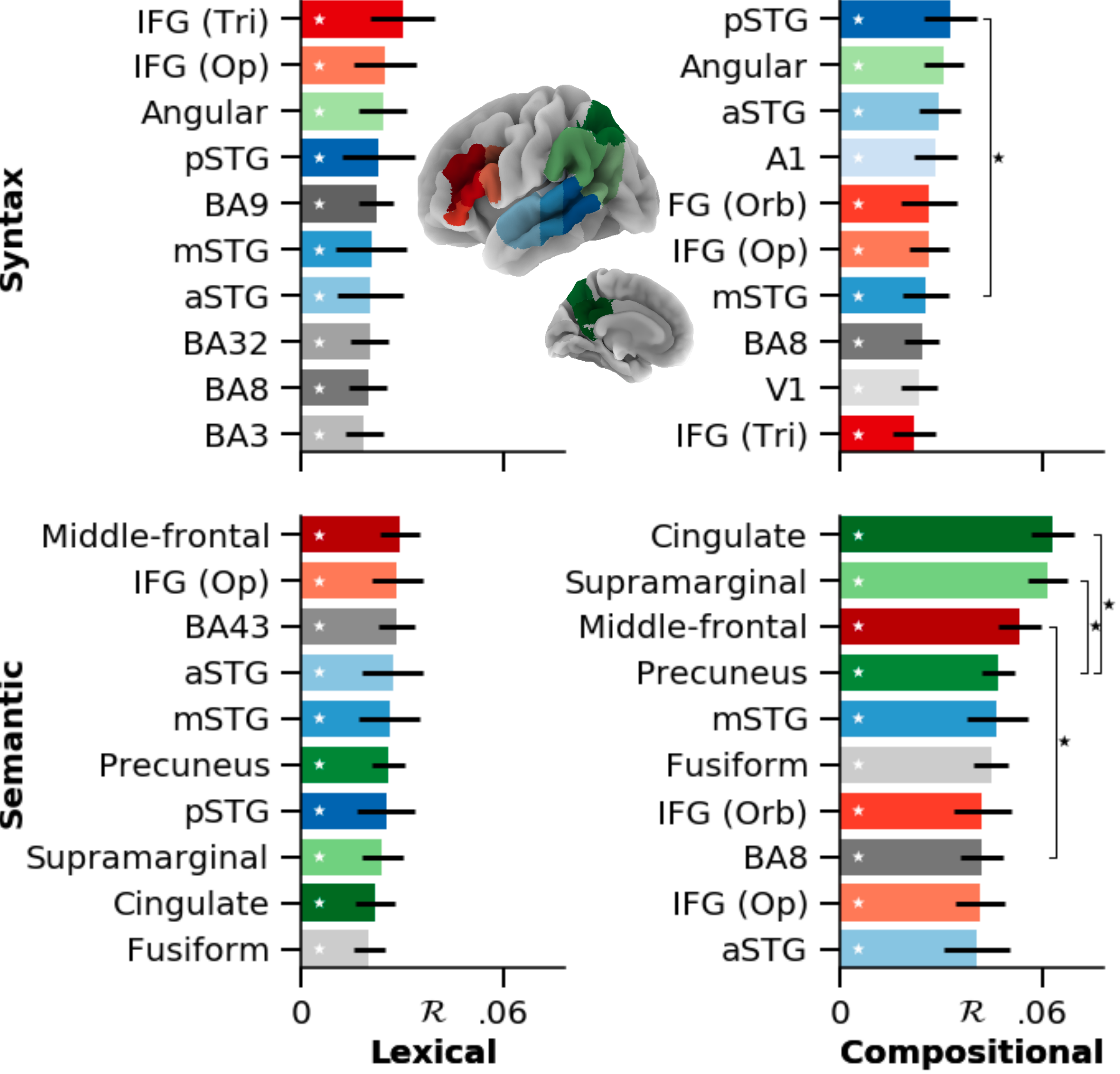}
\caption{
\cameraready{}{
Same as Figure \ref{fig:fig4}.BCEF, with voxel-averaged brain scores (after subtraction of phonological brain scores), for the top ten regions of interest of the left hemisphere (Appendix \ref{SI:brain_parcel}). Error bars are the standard-errors of the mean across the 100 cross-validation folds. Significance (`*') is assessed with a Wilcoxon test across folds, with $\,p<.05$ as a threshold.}}
\vskip -0.25in
\label{fig:rois}
\end{figure}

\section{Experimental Results} \label{sec:res}

\input{results}
\section{Discussion}
\input{discussion}


\section{Acknowledgement}{This work was supported by ANR-17-EURE-0017, the Fyssen Foundation and the Bettencourt Foundation to JRK for his work at PSL.}

\bibliography{main}
\bibliographystyle{icml2021}

\newpage
\quad
\newpage

\appendix

\input{appendix}

\end{document}

%% file: figures.tex
\begin{figure}[ht]
\vskip 0.1in
\centering
\includegraphics[width=\linewidth]{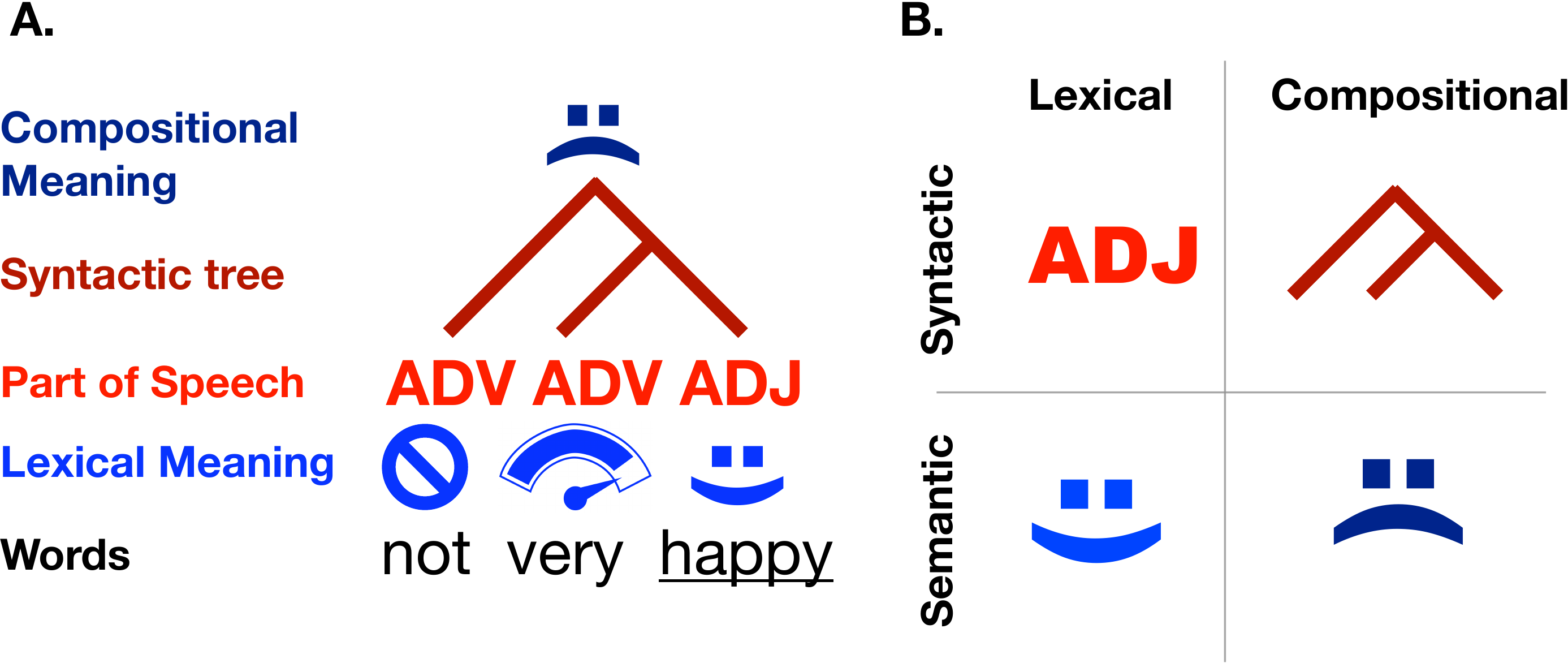}
\caption{\textbf{Taxonomy} 
\textbf{A.} To understand the meaning of a phrase, one must combine the meaning of each word using the rules of syntax. For example, the meaning of the phrase \textsc{not very happy} is (roughly) \textsc{sad}, and can be found by recursively combining the two adverbs and the adjective.
\textbf{B.} \cameraready{We}{Here, we} aim to decompose lexical features (what relates to the word level) from the compositional features (what relates to a combination of words) both for syntactic representations (e.g. part-of-speech versus syntactic tree) and for semantic representations (e.g. the set of word meaning versus the meaning of their combination).
}
\vskip -0.1in
\label{fig:fig1}
\end{figure}

\begin{SCfigure*}
  \caption{\textbf{Method to isolate syntactic representations in GPT-2's word and compositional embeddings.} To isolate the syntactic representations of a sequence of words e.g. $w=$ \textsc{not very happy}, we (1) synthesize sentences with the same syntactic structure as $w$ (e.g \textsc{dimly so true}, \emph{etc.}), then (2) extract the corresponding GPT-2 activations (from layer 9), and finally (3) average these activation vectors across the synthesized sentences. The resulting vector $\overline{X}$ is an approximation of the syntactic representations of $X$ in GPT-2.}
  \includegraphics[width=0.7\textwidth]{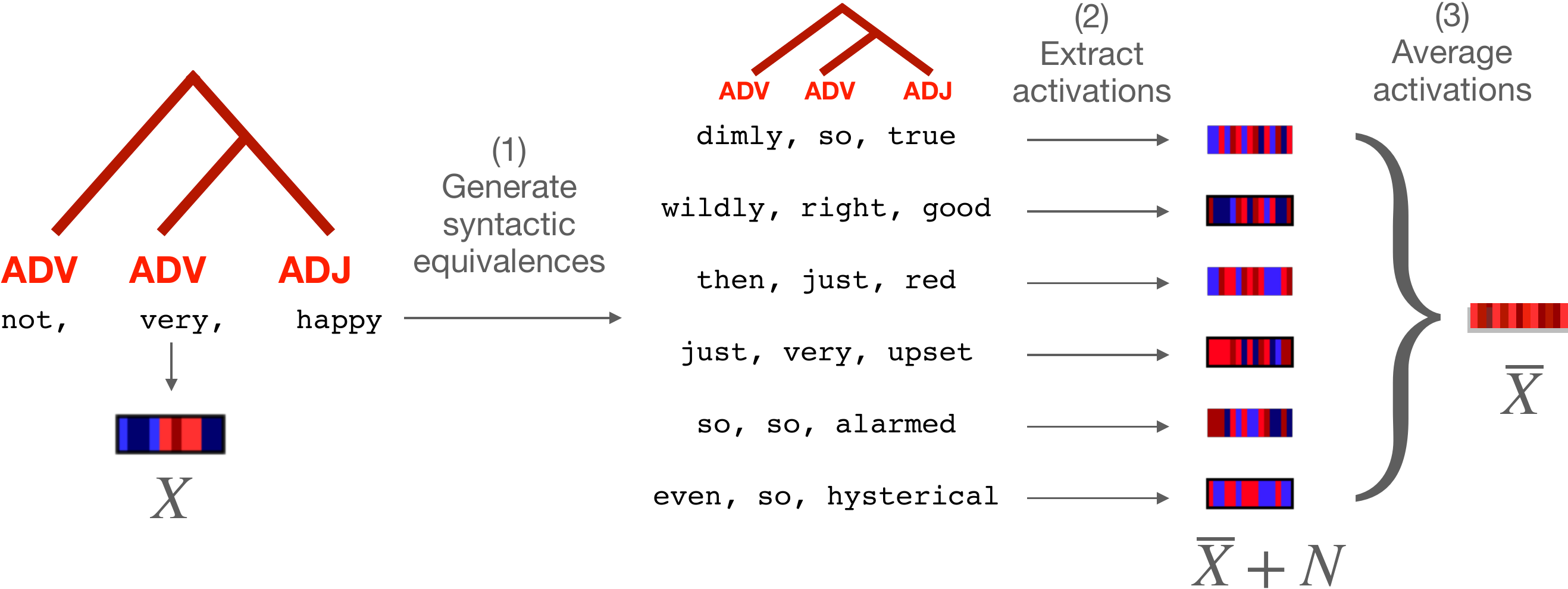}
\label{fig:fig2}
\end{SCfigure*}

%% file: figures2.tex
\begin{figure}[ht]
\vskip 0.1in
\centering
\includegraphics[width=.7\linewidth]{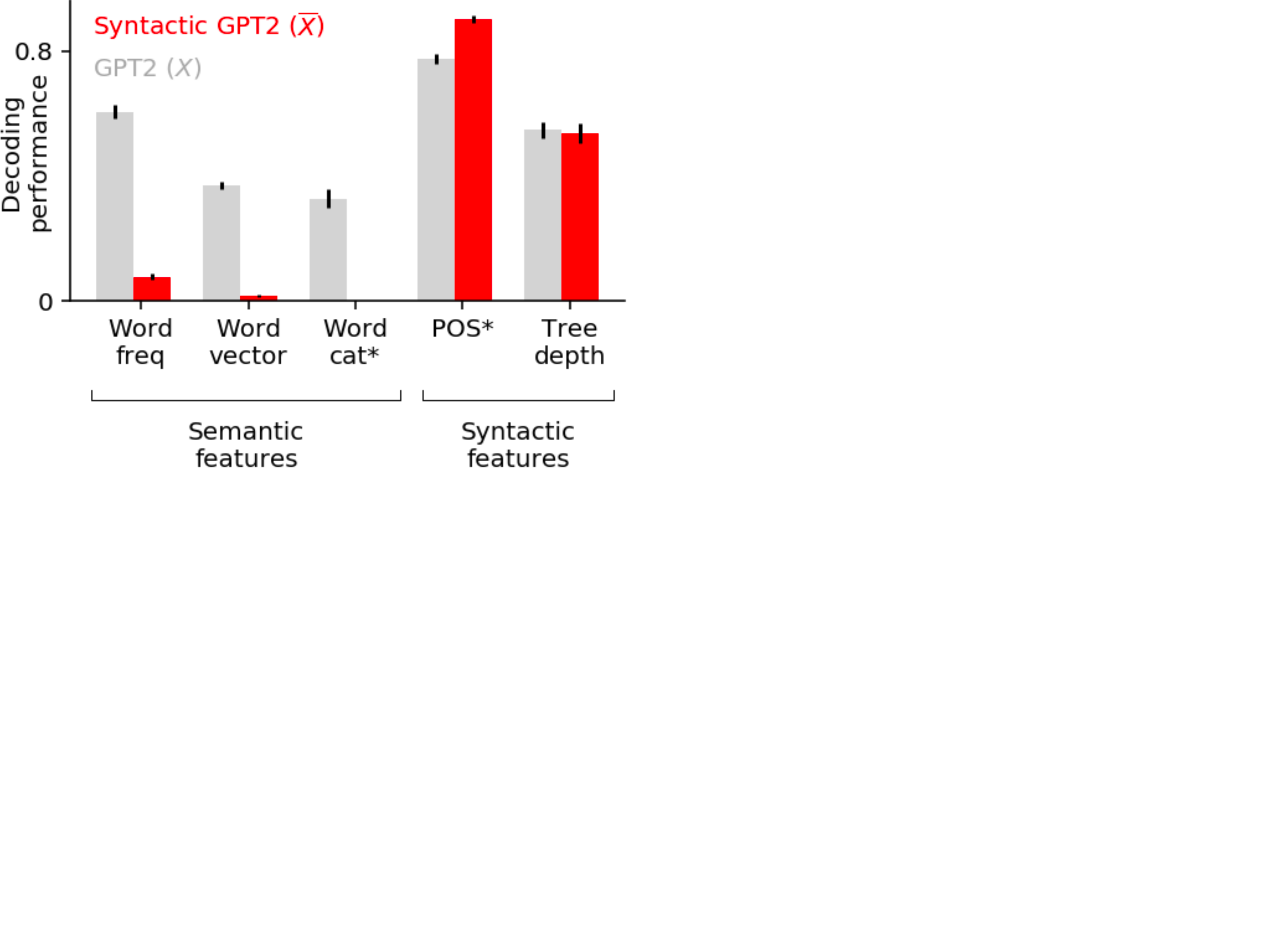}
\caption{\cameraready{}{\textbf{Semantic and syntactic information encoded in $\overline{X}$.} To check that the syntactic embeddings $\overline{X}$ only contain syntactic information, we train a $\ell_2$-regularized linear model to predict three semantic features (frequency, word embeddings and semantic category of content words \cite{binder_toward_2016}) and two syntactic features (part-of-speech and depth of syntactic tree), given the syntactic embedding $\overline{X}$ (red), or the full GPT-2 activations $X$ (grey) (Appendix \ref{SI:probe}). On the y-axis, the decoding performance of the model on left-out data (\textit{adjusted} accuracy for the categorical features marked with a star, $R^2$ for the other continuous features). The chance level is zero.
Semantic features (left) can be decoded from $X$ (grey), but not from $\overline{X}$ (red), while syntactic features (right) can be decoded from both.}}
\vskip -0.1in
\label{fig:probes}
\end{figure}

\begin{figure*}[ht]
\vskip 0.1in
\centering
\includegraphics[width=\linewidth]{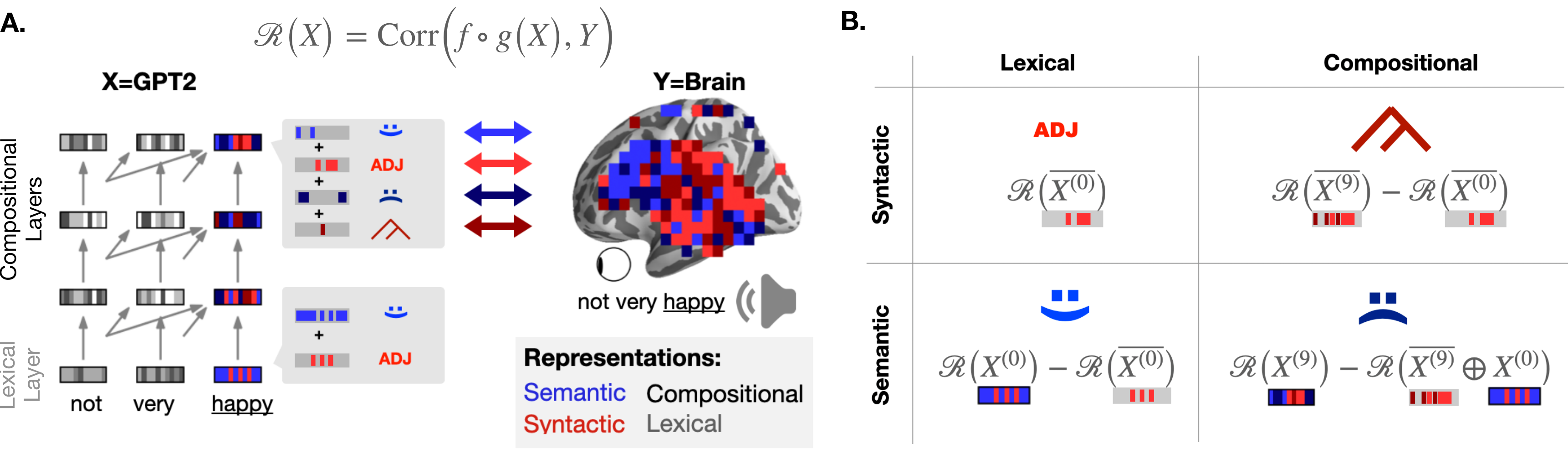}
\caption{ \textbf{Method to decompose the language representations shared between brains and deep language models} \textbf{A.} The human brain and modern language models like GPT-2 both generate \emph{distributed} representations, which are thus difficult to link with the \emph{symbolic} properties of linguistic theories. We introduce a method to decompose the representations of GPT-2, and the corresponding activations $X$ onto the brain activations $Y$, elicited by the same sequence of words (e.g. \textsc{not very happy}) with a spatio-temporal estimator $f \circ g$. This mapping is evaluated through cross-validation, with a Pearson correlation between the predicted and the actual brain signals $\cR$(X).
\textbf{B.} Comparison used to decompose the brain score $\cR(X)$ into the four linguistic components. $X^{(l)}$ refers to the the $l$\textsuperscript{th} layer's activations of GPT-2 input with the sentences heard by the subjects; $\overline{X^{(l)}}$ refers to the average $l$\textsuperscript{th} layer's activations of GPT-2 input with the synthetic sentences with a similar syntax (cf. Figure \ref{fig:fig2}); $\oplus$ indicates a feature concatenation, and '$-$' indicates a subtraction between scores. 
}
\vskip -0.1in
\label{fig:fig3}
\end{figure*}

\begin{figure*}[ht]
\vskip 0.1in
\centering
\includegraphics[width=\linewidth]{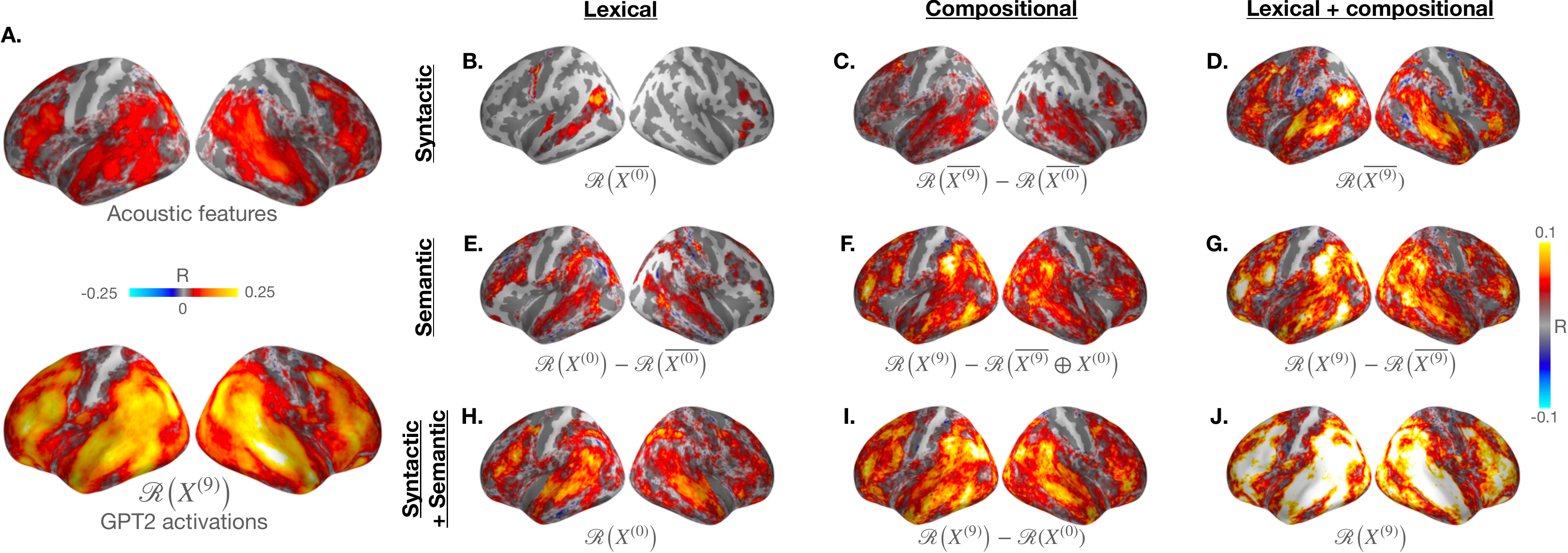}
\caption{ \textbf{Results} Decomposition of the brain scores of 345 subjects listening to narratives into their phonological (A) syntactic (B-D), semantic (E-G), lexical (B-H), compositional (C-I) components and their combinations (ten combinations in total). \textbf{A} Comparison between the brain scores of three phonological features (word rate, phone rate, and phone categories, on the top) and the brain scores of the activations  extracted from the $9^{\mathrm{th}}$ layer of GPT-2, when input with the same narratives (on the bottom). 
\textbf{B-J}. Brain scores decomposed into different sub-processes. To focus on language -- and not low-level speech -- processing, we display the \emph{gain} in brain scores compared to the phonological features. For simplicity, the $\cR$ values reported refers to this gain. 
Brain scores are computed for each fMRI voxel (averaged across subjects), on 100 splits of $\approx$ 2.5\,min of audio stimulus. Non-significant brain regions are not displayed (.05 threshold), as assessed with a two-sided Wilcoxon test across splits,  corrected for multiple comparison across the 75 regions of interest (cf. Section~\ref{SI:brain_parcel}).
}
\vskip -0.1in
\label{fig:fig4}
\end{figure*}

%% file: mapping_new.tex
\cameraready{W}{In the present section, w}e aim to map the activations of two systems $\Psi_1$, a neural network, and $\Psi_2$, the brain, input with the same sequence words $w = (w_1, \dots, w_M)$.  Let $X=\Psi_1(w) \in \mathbb{R}^{M \times d}$ be a vector of $\Psi_1$ activations elicited by $w$ ($M$ vectors of dimension $d$, one per input word), and $Y=\Psi_2(w) \in \mathbb{R}^{N}$ the observable brain response \cameraready{of one voxel,} at each of the $N$ fMRI recorded time sample (a.k.a TR). For simplicity, we consider the analysis for one particular fMRI voxel, the same analysis can be repeated to map $X$ with every voxel in the brain.

To assess the mapping between $X$ and $Y$, we \cameraready{}{use the standard model-based encoding analysis of fMRI signals \citep{huth_natural_2016, yamins2016using,naselaris2011encoding}, and} evaluate a linear spatio- ($f$) temporal ($g$) encoding model trained to predict the $i$\textsuperscript{th} fMRI volume given the network’s activations $X$, on a given interval $I \subset [1\dots N]$:
%
\begin{align} \label{equ_mapping}
\mathcal{R}(X) : f
\mapsto \mathcal{L} 
\bigg( 
    f \circ g(X)_{i \in I}, 
    \overline{(Y_i)}_{i \in I}
    \bigg)
\end{align}


Specifically, given a story $w$ of $M$ words ($w = (w_1,  \dots  , w_M) = (\mathrm{\textsc{the}}, \mathrm{\textsc{cat}}, \mathrm{\textsc{is}}, \mathrm{\textsc{on}}, \mathrm{\textsc{the}}, \mathrm{\textsc{mat}} ,\dots \mathrm{\textsc{end}})$, we first extract the corresponding brain measurements $Y$ of length $N$ time samples. To maximize signal-to-noise ratio, we average the responses across the subjects that listened to that story\cameraready{:}{, and apply the analysis to the average signal $\overline{Y}$.}
\cameraready{
\begin{align*}
\overline{Y} & = \Big(
\frac{1}{|\mathcal{S}|} \sum_{s \in \mathcal{S}} Y_i^{(s)}
\Big)_{i \in I}
\end{align*}

with \(Y^{(s)} = (Y_1^{(s)},  \dots , Y_N^{(s)})\), the $N$ fMRI scans of one subject $s \in \mathcal{S}$, elicited by the story $w$. }{}

The sampling frequency of fMRI is typically lower than word rate. Furthermore, fMRI signals are associated with delayed time responses that can span several seconds. Following others \cite{huth_natural_2016,deniz_representation_2019, shain_fmri_2020}, we \cameraready{thus}{}align the word-times features $X$, of length $M$, to the dynamics of the fMRI signals 
applying a finite impulse response (FIR) model $g$ \cameraready{.
Specifically, for each fMRI time sample $i \in [1 \dots N]$, $g_i$ combines word features within each acquisition interval as follows:
\begin{align*}\label{equ_align}
    g_i : \bbR^{M \times d} & \rightarrow  \bbR^{5d} \\
    u  & \mapsto 
    \big[ 
    \widetilde{ u_i}, 
    \widetilde{ u_{i-1}}, 
    \dots , 
    \widetilde{ u_{i-4}}
    \big] \\
    \widetilde{ u_i} &= \sum_{\substack{m \in \llbracket 1  \dots  M \rrbracket \\ \mathcal{T}(m) = i}} {u_m} \\
\end{align*}
with
\begin{align*}
\mathcal{T} : \llbracket 1  \dots  M \rrbracket & \rightarrow \llbracket 1  \dots   N \rrbracket \\
m & \mapsto i \quad / \quad | t_{y_i} - t_{x_j} | = \min_{k \in \llbracket1 \dots N\rrbracket} | t_{y_k} - t_{x_m} |
\end{align*}

with $\tilde u$ the summed activations of words between successive fMRI time samples, $u$ the five lags of FIR features, $(t_{x_1}, \dots, t_{x_M})$ the timings of the $M$ words onsets, and $(t_{y_1}, \dots, t_{y_N})$ the timings of the $N$ fMRI measurements. }{(cf. Appendix \ref{si:fir})}.

Finally we learn a ``spatial" mapping $f \in \mathbb{R}^{d}$ from the zero-mean unit-variance of $X$ to the zero-mean unit-variance fMRI recordings $Y$ with a $\ell_2$-regularized ``ridge" regression:
\begin{align*}
    \underset{f}{\mathrm{argmin}} \smashoperator[r]{\sum_{i \in I_{\mathrm{train}}}} \Big(\overline{Y_{i}} - f^T g(X)_i\Big)^2 + \lambda ||f||^2
\end{align*}
with $\lambda$ the regularization parameter.
We summarize the mapping with a Pearson correlation score evaluated on left out data:
\begin{equation}\label{equ_corr}
    \mathcal{R} = \mathrm{corr}\Big(f \circ g(X), \overline{Y}\Big) \enspace .
\end{equation}
This correlation score measures the linear mapping between the brain and the activation space $X$. \cameraready{W}{Following others \citep{yamins2016using}, w}e will refer to this score as the \emph{brain score} of \cameraready{an activation space}{the embedding} $X$. 

%% file: decomposing_new.tex
Here, we use the definitions and methods introduced in Section~\ref{taxonomy}, \ref{embed_synax} and \ref{mapping} to decompose the\cameraready{}{ shared} representations of two systems: a deep neural network that encode linguistic properties, and the average brain of 345 subjects listening to narratives.  

To \cameraready{this}{that} end, we (i) compute the activations of the neural language model elicited by the same narratives as the subjects (ii) factorize its activations into linguistic components, (iii) map with supervised learning the factorized components onto brain activity, and finally (iv) decompose the brain activations by evaluating this mapping. 

\cameraready{Deep} Language \cameraready{models}{transformers} are composed of \cameraready{stacked modules called ``}{multiple }layers ($l \in [1\dots L]$), stacked over a (non contextualized) word embedding layer ($l=0$). Each layer can be written as a non-linear system $\Psi^{(l)}$ that transforms a sequence of words $w$ (e.g. \textsc{not}, \textsc{very}, \textsc{happy}) into a vectorial representation of the same length,
\begin{align*}
\Psi^{(l)}: \mathcal{V}^M & \rightarrow \bbR^{M \times d} \\
w & \mapsto \Psi^{(l)}(w) = [\Psi^{(l)}(w)_1, \dots, \Psi^{(l)}(w)_M]
\end{align*}

with $\mathcal{V}$ the set of vocabulary words, $M$ the length of the sequence, and $d$ the dimensionality of the output representation taken at each word. 

We denote $X^{(l)}$ the activations of $\Psi^{(l)}$elicited by $w$, and $\overline{X^{(l)}}$ the syntactic representations extracted from $X^{(l)}$ using the method introduced in Section~\ref{embed_synax}. Following the definitions of Section~\ref{taxonomy}, we can decompose the activations $X$ of $\Psi$ into their:
\begin{itemize}
\item lexical representations: $X^{(0)}$, the word embedding of the network.
\cameraready{}{\item compositional representations: $X^{(l)}, l>\,0$.}
\item syntactic representations: $\overline{X^{(l)}}$, that can be extracted for any layer $l \in [0\dots L]$. The \textit{lexical} syntactic representations $\overline{X^{(0)}}$  is\cameraready{}{roughly} equivalent to the part-of-speech of the word. \textit{Compositional} syntactic representations can be extracted from any layer $l>0$ that encode syntactic information. 
\item  semantic representations: $X^{(l)} - \overline{X^{(l)}}$, as the residuals of syntactic representations. They can be defined at both the lexical $X^{(0)} - \overline{X^{(0)}}$ and compositional levels ($l>0$).
\end{itemize}
\cameraready{}{In practice, to verify that our syntactic embedding ($\overline{X}$) only contains syntax, we evaluate its ability to predict three semantic and two syntactic features (Figure \ref{fig:probes}, Appendix \ref{SI:probe}). 
The results confirm that semantic features can be decoded from $X$ but not from $\overline{X}$, whereas syntactic features can be decoded from both.}

Finally, following Section~\ref{mapping}, we can compute the brain scores of the \cameraready{corresponding}{network's} representations to decompose brain activity into: 
\begin{itemize}
\item lexical representations:  $\cR(X^{(0)})$
\item compositional representations: $\cR(X^{(l)})$, $l>0$.  \cameraready{}{\emph{Strictly} compositional representations are defined as the compositional representations that cannot be explained by lexical features: $\cR(X^{(l)}) - \cR(X^{(0)})$, with $l>0$. For clarity, and except if stated otherwise, we will refer to strictly compositional representations as ``compositional" representations.}
\item syntactic representations: $\cR(\overline{X^{(l)}})$, $l\in [0\dots L]$ 
\item semantic representations: $\cR(X^{(l)}) - \cR(\overline{X^{(l)}})$, i.e. the residual brain scores of syntactic representations, for any layer $l\in [0\dots L]$ 
\end{itemize}

%% file: methods.tex
\section{Experiments \cameraready{}{on the Narratives fMRI Dataset}} \label{expe}

Here, we apply the general\cameraready{statistical}{} method described \cameraready{above}{in Section~\ref{embed_synax}, \ref{mapping} and \ref{decomposing}} to decompose the activations of two nonlinear systems, GPT-2 ($\Psi_1$) and the brain activity of 345 subjects listening to narratives ($\Psi_2$). 
\paragraph{Functional MRI dataset.}
We analyze the \cameraready{publicly available} ``Narratives" \cameraready{}{public} dataset \cite{nastase_narratives_2020}\cameraready{}{, which contains} \cameraready{comprised of} the fMRI measurements of 345 unique subjects listening to narratives. The narratives consist of 27 English spoken stories, ranging from $\approx 3$ minutes to $\approx 56$ minutes, for a total of $\approx$ 4.6 hours of unique stimuli. The original paper included two fMRI preprocessing pipelines, one with spatial smoothing and the other without. All our analyses are tested on the unsmoothed fMRI. As suggested in the original paper, we exclude (story, subject) pairs because of noisy fMRI recordings or missing transcripts, resulting in 617 unique (story, subject) pairs in total and  $\approx$ 4 hours of unique audio stimuli.

\paragraph{Phonological features.}
To focus on lexical and supra-lexical language processing -- as opposed to low-level speech processing, we extract three potential \cameraready{phonological}{sets of} confounds: the phone rate (the number of phones between two fMRI measurements, of dimension 1), the word rate (the number of words between two fMRI measurements) and the concatenation of the phoneme, stress and tone of the words in the stimulus. 
For each story,
a phoneme-level transcript was provided in the Narratives database thanks to Gentle\footnote{https://github.com/lowerquality/gentle}, a forced-alignment algorithm. 
\cameraready{Using} Gentle annotations \cameraready{}{led to }\cameraready{,} 117 unique \cameraready{phones}{categories (with unique phone, stress and tone)}\cameraready{ were found in total}, resulting  in a one-hot encoded feature of the same dimension. 


\paragraph{\cameraready{Neural language models activations}{Language model features}.}
GPT-2 is a high-performing causal (i.e. left to right) language model trained to predict a word given its previous context \cite{radford_language_nodate}, and known to generate brain-like representations \citep{goldstein2021thinking,caucheteux2020language,affolter2020brain2word, schrimpf_artificial_2020,caucheteux2021gpt}. It is comprised of 12 Transformer (contextual) layers ($l \in [1\dots12]$) stacked over a (non-contextual) embedding layer ($l=0$), each of dimensionality 768, with 1.5 billion parameters in total. We used the pretrained version of GPT-2 from Huggingface \cite{wolf-etal-2020-transformers}, trained on a dataset of 8 million web pages. 
\cameraready{In practice, to extract the activations elicited by the 27 stories, we proceed as follows: we first format and lower cased the texts (replacing special punctuation marks such as ``--" and duplicated marks ``?." by dots), then apply the tokenizer provided by Huggingface to convert the transcript into either word-level or sub-word-level tokens called ``Byte Pair Encoding" (BPE) \cite{sennrich_neural_2016}. Here, more than 99.5\% of BPE-level tokens were complete words. The tokens are then split into sections of 256 tokens (this length is constrained by GPT-2's architecture) and input to the model one story at a time. The activations of each layer are finally extracted, resulting in 12 vectors of 768 activations for each token of each story transcript (i.e. one for each of the 12 layers).}{In practice, the 27 stories are pre-processed, tokenized and input to the model (Appendix \ref{SI:other_transformers}).
The activations of each GPT-2 layer are extracted, resulting in 12 vectors of 768 activations for each token of each story transcript. 
For comparison, we also study five other transformers: BERT \cite{devlin_bert_2019}, XLnet \cite{yang_xlnet_2020}, Roberta \cite{liu_roberta_2019}, AlBert \cite{lan_albert_2020} and DistilGPT-2 (a smaller version of GPT-2) and recover similar -- although lower -- brain scores (Appendix \ref{SI:other_transformers}).}
%
%
\paragraph{
\cameraready{}{Extracting syntactic representations from GPT-2}
.}
To isolate the syntactic representations of GPT-2 
, we \cameraready{generate}{synthesize}, for each \cameraready{story $w$}{sentence of each story}, $k=10$ \cameraready{texts}{sentences} with the same syntactic structures (Figure \ref{fig:fig2}). \cameraready{}{We ensure in supplementary analyses that (i) the $k$ synthetic sentences do \textit{not} include the target sentence and (ii) these syntactic embeddings ($\overline{\Psi_k}$) lead to stable representations of syntax (Appendix \ref{SI:convergence})}. To this end, we proceed as follows:

\begin{itemize}
    \item The transcript is formatted, split into sentences and tokenized using the large English tokenizer provided by spaCy \cite{spacy} (cf. Appendix \ref{SI:other_transformers}).
    \item Then, we use Supar, a state-of-the art dependency parser \cite{zhang-etal-2020-efficient} to extract the dependency structure of each sentence and the part-of-speech.
    \item For each \cameraready{}{target} word of \cameraready{a}{each} sentence \cameraready{}{of the Narratives dataset}, we sample\cameraready{}{, from a $\approx$ 58,000 word corpus, consisting of Wikipedia combined with Narratives' transcripts, up to} \cameraready{from $k'=\,100$} to $k'=\,1,000$ words \cameraready{of}{that have} the same part-of-speech and dependency tags \cameraready{}{(e.g. \textsc{cat: noun, singular, subject of})}. At this stage, $k'$ versions of the \cameraready{initial}{target Narratives} transcripts \cameraready{were generated}{are synthesized}. 
    \item The \cameraready{reconstructed}{synthesized} sentences are not always grammatically correct. Thus, we automatically correct \cameraready{incorrect} the sentences with Gector \cite{omelianchuk_gector_2020}, and filter out the sentences that do not have the same length or part-of speech as the \cameraready{initial}{target} sentence \cameraready{}{in the Narratives corpus}.
    
    \item Some of the generated sentences may end up with a distinct syntactic tree than the original sentence, because semantics can disambiguate syntax (e.g. \textsc{I shot an elephant in my pyjamas}). To assess the syntactic similarity between the original and the generated sentences, we compute\cameraready{d}{, from} their respective syntactic trees, \cameraready{and computed, }the Pearson correlation between the words' pairwise distances, following \cite{manning_emergent_2020}'s method. Then, we select\cameraready{ed}{} the sentences whose syntactic trees are the most similar. 95\% of the generated sentences \cameraready{had}{have} a syntactic tree that correlate\cameraready{d}{s} with the \cameraready{original}{tree of the target} sentence\cameraready{'s}{} above R=90\%.
\end{itemize}


\paragraph{\cameraready{}{Mapping embeddings onto onto the fMRI signals.}}
As described in equation \eqref{equ_mapping}, we evaluate the mapping between \cameraready{}{a set of modeling features $X$} \cameraready{the activations $X^{(l)} \in \bbR^{M\times d_x} $ extracted from the layer $l$ of GPT-2,}{} and the fMRI signals $Y \in \bbR^{N\times d_y}$ by fitting a linear spatio- ($f$) temporal ($g$) encoding model. $f \circ g$ was fitted on $I_{\mathrm{train}}=$ 99\% of the dataset, and evaluated on $I_{\mathrm{test}}=$ 1\% of the left out-data (2.5 min of audio)\cameraready{ using $\mathcal{L}$,}{}. \cameraready{}{We evaluate the quality of this mapping with} a Pearson R correlation between predicted and actual brain signals on $I_{\mathrm{test}}$. \cameraready{In practice}{Specifically}, we use\cameraready{d}{} the linear ridge regression from scikit-learn \cite{scikit-learn}, with penalization parameters chosen among 10 values log-spaced between $10^{-1}$ and $10^8$ and $g$ was a finite impulse response (FIR) model with 5 delays, following \cite{huth_natural_2016}. $X$ and $Y$ \cameraready{were}{are} normalized (mean=0, std=1) across scans for each story, using a robust scaler clipping below and above the $0.01^{st}$ and $99.99^{th}$ percentiles, respectively. We repeat\cameraready{ed}{} the procedure 100 times with a 100-fold cross-validation,  using scikit-learn `KFold' without shuffling \citep{scikit-learn}.



%
\paragraph{Statistical significance.}
We assess the significance of our results across test folds ($k=100$). To this end, we first average the brain scores within each brain region, as defined by the Destrieux Atlas parcellation \cite{destrieux_automatic_2010}. Then, we apply a Wilcoxon two-sided signed-rank test across folds to evaluate whether this average brain score is significantly different from zero. The p-values of the 75 brain regions were corrected for multiple comparison using a False Discovery Rate, (Benjamini/Hochberg) as implemented in MNE-Python \cite{gramfort_meg_2013}. Non-significant p-values (\cameraready{.05 threshold}{$p \geq \,.05$}) are masked in Figure \ref{fig:fig4}.

%% file: results.tex


%
\paragraph{Phonological features.}
\cameraready{First, t}{T}o isolate the sublexical speech representations, we compute the brain scores using a concatenation of three sets of features, \emph{i.e.}, word rate, phone rate, and phone categories. These \cameraready{}{sublexical} features lead to significant brain scores across the expected language networks and mainly peak within the bilateral superior temporal lobe, the temporo-parietal junction, the lateral intra-parietal sulcus, the infero-frontal cortex (IFG) as well as in the right motor cortex (Figure \ref{fig:fig4}A and \ref{fig:rois}).

To \cameraready{strictly focus on}{isolate} lexical and compositional representations, we \cameraready{now} focus \cameraready{}{the next analyses} on the \emph{gain} in brain scores obtained over those of sublexical features (\emph{i.e.} to the increase of brain scores obtained with each feature set, as compared to the scores obtained with phonological features). For simplicity, the $\cR$ scores reported in Figure \ref{fig:fig4}, \ref{fig:rois} and \cameraready{}{in the text below }refer to this gain.

The brain scores corresponding to the lexical ($\cR(X^{(0)})$), compositional ($\cR(X^{(9)})$), syntactic ($\cR(\overline{X^{(9)}})$ and semantic representations ($\cR(X^{(9)}) - \cR(\overline{X^{(9)}})$) of the ninth layer of GPT-2 are displayed in figures \ref{fig:fig4}\cameraready{We focus on GPT-2's ninth layer because this intermediate layers have been shown to better encode syntactic properties than input and output layers.}{ and \ref{fig:rois}} (non-significant scores after correction for multiple comparisons across regions are masked).

\paragraph{Lexical features.}
The lexical representations of the brain \cameraready{can be}{have been repeatedly} investigated through the lens of a word-embedding \citep{mitchell2008predicting,huth_natural_2016,toneva2019interpreting,schrimpf_artificial_2020,caucheteux2020language}.
\cameraready{}{Here, we replicate these analyses: }GPT-2's word embedding $X^{(0)}$ \cameraready{results in}{leads to} lexical brain scores significantly higher than sublexical features' in most of the language network, \emph{i.e.} in the bilateral superior temporal lobe and the infero-frontal cortex (Figure \ref{fig:fig4}H). 
%
\paragraph{Lexical syntax.}
Do these brain scores result from semantic and/or syntactic representations? To tackle this issue, we compute brain scores from the word embeddings ($\overline{X^{(0)}}$) input with \cameraready{the shuffled}{synthesized and syntactically-matched} sentences: \emph{i.e.} word sequences sharing the same syntax as the \cameraready{original}{target} sentence \cameraready{}{in the original Narratives corpus} (Figure \ref{fig:fig4}B). The results reveal significant brain scores (\cameraready{as compared to}{\emph{i.e.} higher than} sublexical ones) \cameraready{mainly in the superior temporal sulcus, the left planum temporale, as well as in the infero- and orbito-frontal cortex}{in a distributed network including the infero-frontal cortex, the angular gyrus and the posterior superior temporal gyrus (Figure \ref{fig:rois})}.

\paragraph{Lexical semantics.}
To identify the representations of lexical semantics, we compare the brain score obtained with the word embedding to those obtained with the embedding of lexical syntax ($\cR(X^{(0)}) - \cR(\overline{X^{(0)}})$ in Figure \ref{fig:fig4}E). The resulting brain scores \cameraready{were}{are} significant mainly in the left hemisphere, and peak\cameraready{ed}{} in the superior temporal gyrus, the infero-frontal cortex\cameraready{and in the supra-marginal cortex}{} as well as in the precuneus and the tranverse temporal gyrus. These results are more modest than we anticipated given past work \citep{huth_natural_2016}.

\paragraph{Compositional representations.}
Recent studies have shown that the contextual (\emph{i.e.} deep) layers of language models better predict brain activity than word embedding \citep{jain2018incorporating,jat2019relating,toneva2019interpreting,caucheteux2020language}.
We replicate this result with a representative contextual layer of GPT-2 (layer 9 out of 12, Figure \ref{fig:fig4}J):  $\cR(X^{(9)})$ almost doubles the brain scores obtained with the word embedding $\cR(X^{(0)})$ in the bilateral temporal, infero-frontal and infero-parietal cortices. 

\paragraph{Compositional syntax.}
Do these gains in brain score reflect compositional semantics and/or compositional syntax? To tackle this issue, we compare the brain scores obtained with the ninth layer of GPT-2 input with the \cameraready{shuffled}{syntax-matched synthesized} sentences $\cR(\overline{X^{(9)})}$, to the the brain scores obtained with the first layer of GPT-2, input with those same \cameraready{shuffled}{synthesized} sentences  $\cR(\overline{X^{(0})}$.
The results show that the representations of compositional syntax are distributed over the bilateral temporal and infero-frontal cortices\cameraready{pars opercularis}{}, and actually extend to a relatively large set of brain areas\cameraready{}{ (Figure \ref{fig:fig4}C-D)}.
Overall, these results, although correlational, thus favor a distributed \citep{fedorenko2012lexical} rather than a modular \citep{pallier2011cortical,friederici2000segregating} view of syntax: both lexical and compositional syntactic effects do not appear to be confined within a single brain area.

\paragraph{Compositional semantics.}
Finally, we estimate the brain representations of compositional semantics by comparing the brain scores obtained with the syntactic representations $\cR(\overline{X^{(9}})$ to those obtained with the ``normal" activations $\cR(X^{(9})$, \emph{i.e.} GPT-2's activations obtained with the same sentences as subjects heard. Again, the resulting effects proved to be remarkably distributed, and peaked in the \cameraready{}{cingulate, }supramarginal, and \cameraready{infero}{middle}-frontal cortex \cameraready{pars triangularis}{} (Figure \ref{fig:fig4}G). These brain scores appear to result from strictly compositional semantics: these effects remain significant even when we subtract away the contribution of lexical semantics (Figure \ref{fig:fig4}E and \ref{fig:rois}).
\begin{figure}[ht!]
\vskip 0.1in
\centering
\includegraphics[width=.8\linewidth]{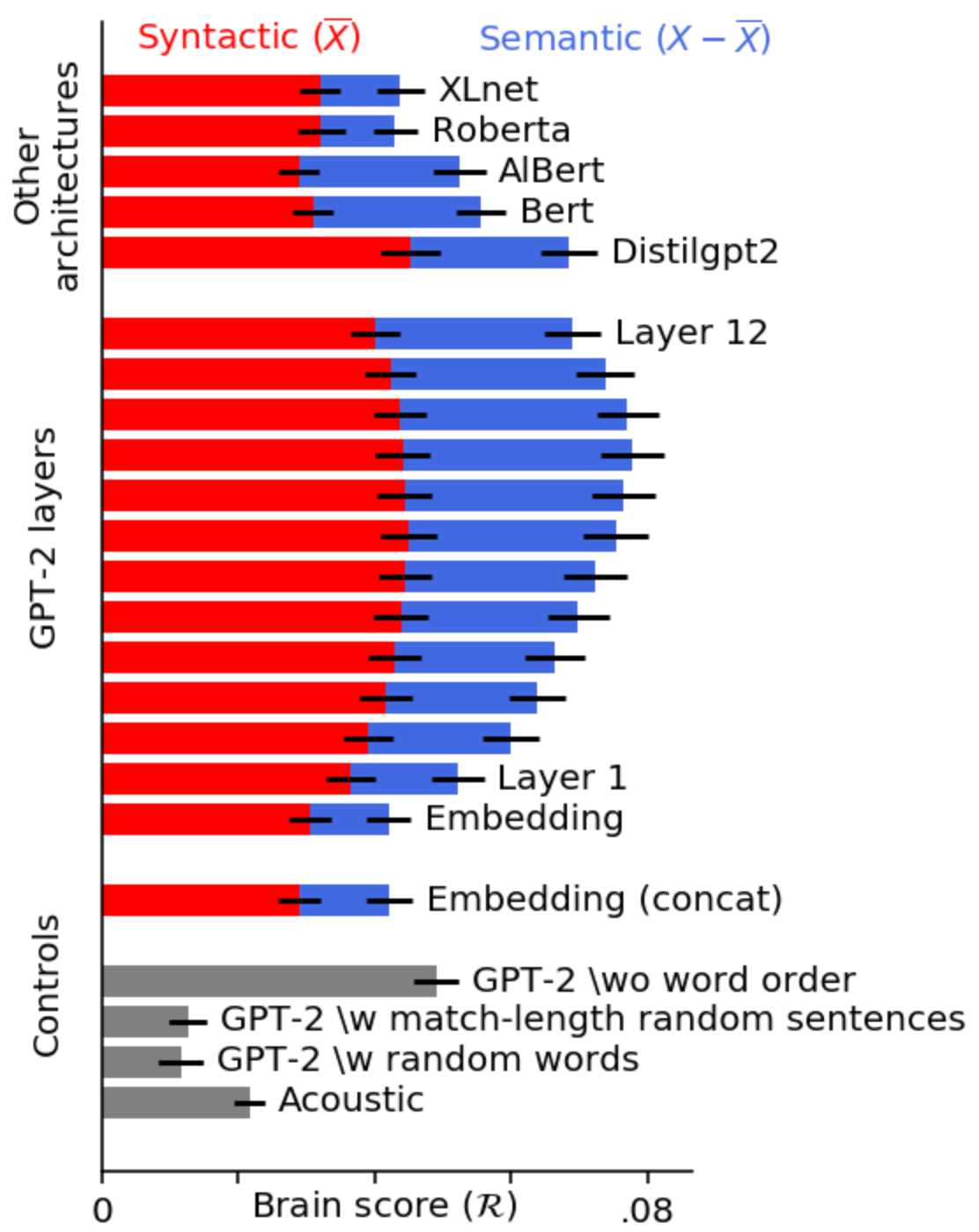}
\caption{
\cameraready{}{
\textbf{Generalisation to other layers and architectures} In red, the brain scores of the syntactic embeddings ($\mathcal{R}(\bar X)$) built out of GPT-2 layers (from the word embedding to layer 12), and the middle layer of five transformer architectures (top, cf. Appendix \ref{SI:other_transformers}, $l=2/3 \times n_{\mathrm{layers}}$). In blue, the residuals of syntax ($\mathcal{R}(X) - \mathcal{R}(\bar X)$) in the brain. Bottom, the brain scores of i) acoustic features (the concatenation of word rate, phoneme rate, phoneme stress and tone), GPT-2 activations induced ii) by random words sampled in the stimulus, iii) by sentences randomly sampled from Wikipedia, matching in length with the sentences of the stimulus, iv) by the actual sentences of stimulus, but with random word order in each sentence (Appendix \ref{SI:controls}.})}
\vskip -0.1in
\label{fig:generalisation}
\end{figure}

\paragraph{Control 1: low-level linguistic properties.}
\cameraready{}{Do the syntactic representations evidenced above simply capture the length of sentences? To address this issue, we input the above analyses with i) random words sequences (\emph{i.e.} non grammatical) and ii) random but well-formed sentences that have the same length as those of the Narratives corpus.
The results show that neither of these two embeddings match the brain scores obtained with syntactic and/or semantic representations (Figure \ref{fig:generalisation}). Similarly, using the GPT-2 activations elicited by the sentences of the Narratives after a random word permutation leads to lower brain scores than our original analyses. Together, these results confirm that our decomposition of syntactic and semantic representations in the brain cannot be reduced to simplistic representations like bags of words and/or sentence length.}


%
\paragraph{Control 2: generalisation to other layers and architectures.} 
\cameraready{}{The above results are obtained using the ninth layer of GPT-2.
We chose to study this model and this layer, because a) GPT-2, like the brain, processes words in a \emph{causal} way, b) it is known to best predict brain responses \cite{schrimpf_artificial_2020,caucheteux2021gpt}, c) its middle layers best encode complex semantic and syntactic properties \cite{jawahar_what_2019,manning_emergent_2020}. To test the generality of our study, we apply the same analyses to five other language transformers as well as to all of the layers of GPT-2 (Figure \ref{fig:generalisation}). The results generalize to each layer of GPT-2, and peak around layer 9. The five other transformers (for their middle layer $l=2/3 \times n_{\mathrm{layers}}$) result in similar, although significantly lower brain scores (Appendix \ref{SI:other_transformers}).
}

%% file: discussion.tex

In the present study, we introduce a simple taxonomy and its associated method to decompose the distributed representations of language in brains and deep language models.



Our taxonomy capitalizes on classic linguistic proposals \citep{lycan2018philosophy, givon2001syntax,chomsky2014minimalist} to offer precise definitions of lexicality, compositionality, syntax and semantics\cameraready{.}{, which operate on \emph{distributed} representations. Our results show that these four sets of linguistic features, typically theorized in terms of discrete symbols, can be, as long predicted \citep{smolensky_tensor_1990}, investigated in artificial and biological neural networks}. 

\cameraready{These definitions, however, are far from perfect}{The present definitions remain imperfect}. First, compositionality is often associated with specific properties that are not presently considered (e.g. systematicity and generalisation \citep{szabo2004compositionality,hupkes2019compositionality,baroni2020linguistic}). Furthermore, we here define semantics as the \emph{residual} representations of any text embedding once syntactic representations have been removed. This proposal is very coarse: semantics is generally defined as the study of meaning (which is itself not \cameraready{very}{} easy to define). Yet, some language features like emotional value and textual style may arguably not ``mean" anything, in that they do not necessarily refer to a state of the world \cameraready{}{and yet would be categorized as semantics according to our proposed taxonomy}. In spite of \cameraready{this limit}{these limits}, the advantage of our framework is that it makes \cameraready{}{simple, }precise and quantifiable predictions to investigate distributed linguistic representations \cameraready{}{in the human brain. Furthermore, the present framework is particularly versatile in that i) it can, in principle accommodate any natural sentences and ii) its conclusions can be refined with the development of better and/or more biologically-plausible models of language}.
 
The present study follows suit with past research on naturalistic \cameraready{}{and thus} poorly-controlled linguistic stimuli \citep{mesgarani2014phonetic,huth_natural_2016,brennan2016naturalistic,brennan2019hierarchical,stehwien2020little,gwilliams2020neural}.\cameraready{}{ While we replicate previous neuroscientific findings regarding lexical semantics (Figure \ref{fig:fig4}E) \cite{huth_natural_2016} and lexical \emph{vs} compositional processing in the brain (Figure \ref{fig:fig4}.H,J) \cite{toneva2019interpreting, schrimpf_artificial_2020, goldstein2021thinking}, \cameraready{our other neuroscientific results are novel }{our systematic decomposition of language representations brings new light on the brain bases of syntax (Figure \ref{fig:fig4}.BCDFG)}}. In addition, \cameraready{this}{our} approach diverges with and complements previous practices, consisting of carefully designed stimuli, typically matched for word length, word frequency \citep{kutas1980reading} and/or constituent size \citep{pallier2011cortical,ding2016cortical}, which becomes exponentially difficult when the number of variables to control increases \citep{hamilton2020revolution}. This change of \cameraready{practice}{paradigm} has been empowered by the rise of high-performing language models: previous research lacked a method to make single trial/single sentence predictions and could thus only compare the average activations across blocks of similarly constructed sentences. By contrast, modern language models offer the possibility to predict the representations of individual words and sentences \citep{hale2018finding,toneva2019interpreting,caucheteux2020language, schrimpf_artificial_2020,heilbron2020hierarchy}. Consequently, carefully-controlled experimental designs can now be relaxed to naturalistic settings, and allow one to refine her tests and hypotheses without having to conduct new (and arguably artificial\cameraready{e.g. Jabberwocky \citep{pallier2011cortical,fedorenko2012lexical}}) experiments. 

The main drawback of such an uncontrolled setting is undoubtedly signal-to-noise ratio: like any bias/variance trade-off, relaxing the set of hypotheses that one can test in a given dataset reduces the probability of a successful \cameraready{test}{finding}. To accommodate this issue, we here opted to analyze the average brain signal across subjects. Even then, \cameraready{prediction}{brain} scores remain far from 100\%. \cameraready{Furthermore,}{Given that} the brain bases of language\cameraready{ processing} are notoriously variable across individuals \citep{fedorenko2010new} \cameraready{. F}{f}uture works \cameraready{thus}{} remain necessary to \cameraready{take into}{better account for} the functional and anatomical variability across subjects.

Thanks to machine learning, our method sheds new light on the neural bases of language in general, and of syntactic processes in particular. First, it supplements previous work on the neural basis of lexical \citep{friederici2000segregating,mitchell2008predicting}
and compositional representations of language \citep{pallier2011cortical,nelson_neurophysiological_2017,fedorenko2012lexical,brennan2017meg}: syntactic processes, in particular, appear to be linked to a remarkably wide-spread \emph{distribution} of activation in the language networks. This result favours a distributed \citep{fedorenko2012lexical} as opposed to a modular \citep{pallier2011cortical,friederici2000segregating} view of syntactic processes. Second, our study highlights the remarkably-large recruitment of compositional semantics -- an observation that strengthens and extends what had already been reported at the lexical level \citep{huth_natural_2016}. Overall, these results thus reinforce the idea that speech comprehension \cameraready{likely}{} results from the coordination of a huge cortical network. While its \cameraready{rules}{functional principles} remain \cameraready{unknown}{largely unexplored}, the \cameraready{functional}{} similarity between the \cameraready{}{human }brain and deep \cameraready{networks}{language models} offers a new and powerful mean to understand the laws of language.

%% file: appendix.tex
\begin{center}
    \Large
    \textbf{Appendix}
\end{center}
\medskip

\section{Deep Neural Networks' Activations}\label{SI:other_transformers}

\paragraph{Pre-trained tansformers}
In Section \ref{expe}, we extract the activations of GPT-2 \cite{radford_language_nodate} and five transformer architectures: BERT \cite{devlin_bert_2019}, XLnet \cite{yang_xlnet_2020}, Roberta \cite{liu_roberta_2019}, AlBert \cite{lan_albert_2020} and DistilGPT-2. We use the pre-trained models from Huggingface \cite{wolf-etal-2020-transformers}: `bert-base-cased', `xlnet-base-cased', `roberta-base', `albert-base-v1', and `distilGPT-2' respectively. In Figure \ref{fig:generalisation}, we focus on one middle layer of these transformers ($l = n_{\mathrm{layers}} \times 2/3$), because it has shown to best encode brain activity \cite{caucheteux2020language} and to encode relevant linguistic properties \cite{manning_emergent_2020, jawahar_what_2019}. 

\paragraph{Text formatting and tokenization}
To extract the activations elicited by one story, we proceed as follows: we first format and lower case the text (replacing special punctuation marks such as ``--" and duplicated marks ``?." by dots), then apply the tokenizer provided by Huggingface \cite{wolf-etal-2020-transformers} to convert the transcript into either word-level or sub-word-level tokens called ``Byte Pair Encoding" (BPE) \cite{sennrich_neural_2016}. Here, more than 99.5\% of BPE-level tokens were complete words. The tokens are then split into sections of 256 tokens (this length is constrained by GPT-2's architecture) and input to the deep network one story at a time. The activations of each layer are finally extracted, resulting in $n_{\mathrm{layers}}$ vectors of 768 activations for each token of each story transcript. In the 0.5\% case where BPE are not complete words, BPE-features are summed between successive words, to obtain $n_{\mathrm{layers}}$ vectors per word per story.

\section{Convergence of the Method to Build $\overline{X}$} \label{SI:convergence}

\begin{figure}[ht]
\vskip 0.1in
\centering
\includegraphics[width=\linewidth]{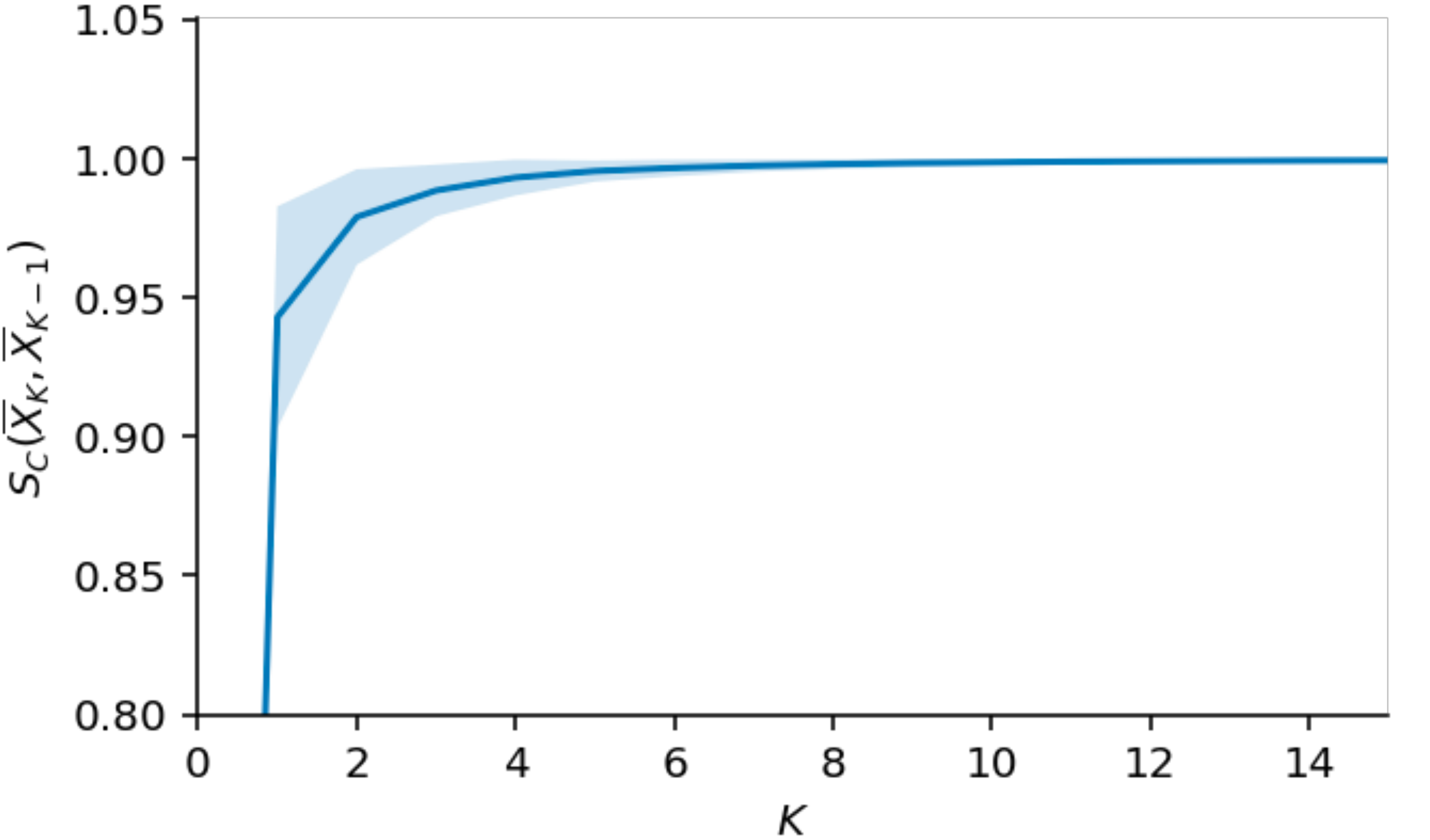}
\caption{\textbf{Convergence of the method to build syntactic embeddings.} Cosine similarity $S_C$ between the syntactic component $\overline{X}$ of GPT-2 activations induced by a sequence $w$, when computed with $K$ and $K-1$ syntactically equivalent sequences. The syntactic embeddings $\overline{X}_K$ and $\overline{X}_{K-1}$ are computed for 100 Wikipedia sentences ($\approx 2,800$ words), and the similarity scores are averaged across embeddings. In shaded, the 95\% confidence interval across embeddings.}
\vskip -0.1in
\label{fig:si_convergence}
\end{figure}

In Section \ref{embed_synax} and \ref{decomposing}, we compute the syntactic component $\overline{X}$ of GPT-2 activations $X$ elicited by a sentence $w$. $\overline{X}$ is approximated by $\overline{X_k}$, the average activations across $k$ sentences with the same syntax as $w$. Here, we sample $k=10$ sentences. We check in Figure \ref{fig:si_convergence} that the method has converged before $k=10$. We compute the cosine similarity between $\overline{X}_k$ and $\overline{X}_{k-1}$ for $k$ between 1 and 15. The syntactic embeddings stabilize with at least eight sampled sentences. 

\section{Evaluating the Level of Semantic and Syntactic Information in $\overline{X}$}\label{SI:probe}

In Section \ref{decomposing} and Figure \ref{fig:probes}, we check that the syntactic embedding $\overline{X}$ extracted from GPT-2 only contains syntax. To this aim, we evaluate the ability of a linear decoder to predict two syntactic features and three semantic features from  $\overline{X}$.

\paragraph{Semantic and syntactic features} 

The two syntactic features derived from the stimulus are:
\begin{itemize}
    \item The part-of-speech of the words (categorical feature), as defined by Spacy tags \cite{spacy}. 
    \item The depth of the syntactic tree (continuous feature). The syntactic tree is extracted with the state-of-the-art Supar dependency parser \cite{zhang-etal-2020-efficient}.
\end{itemize}

The three semantic features are only computed for verbs, nouns and adjectives (as defined by Spacy part-of-speech tags) and are the followings:
\begin{itemize}
    \item Word frequency (labeled as `Word freq' in Figure \ref{fig:probes}, continuous feature). We use the `zipf\_frequency' from the wordfreq\footnote{https://pypi.org/project/wordfreq/} python library.
    \item Word embedding (continuous feature), computed using the pre-trained  model from Spacy \cite{spacy} (`en\_core\_web\_lg', 300 dimensions).
    \item Semantic category (categorical feature). We used the 47 semantic categories\footnote{Categories are: abstract, action, animal, auditory, body, building, cognitive, construct, creative, device, distant, document, electronic, emotion, emotional, entity, event, food, furniture, general, geological, group, human, instrument, locative, mental, miscellaneous, multimodal, object, part, perceptual, period, physical, place, plant, property, social, somatosensory, sound, spatial, state, temporal, time, tool, vehicle, visual, weather}. Categories are not available for all the 2,800 Wikipedia words studied here. Thus, we first train a linear model (scikit-learn `RidgeCVClassifier') to predict the semantic category of the 535 labeled words used in \cite{binder_toward_2016}, given their Spacy word embedding (300 dimensions). We then label the 2,800 Wikipedia words using the semantic category predicted by the classifier.
\end{itemize}

\paragraph{Linear decoder}
To evaluate the ability of a linear decoder to predict the five linguistic features from $\overline{X}$, we:
\begin{itemize}
    \item Build syntactic embeddings $\overline{X}$ for 100 Wikipedia sentences ($\approx 2,800$ words), following Section \ref{embed_synax}, using the ninth layer of GPT-2. 
    \item Build the three semantic and two syntactic features described above from the 2,800 Wikipedia words Wikipedia words.
    \item Fit a $\ell_2-$regularized linear model to predict the five features given the syntactic embeddings. We use the `RidgeCV' regressor (resp. `RidgeClassifierCV' classifier) from scikit-learn \cite{scikit-learn} to predict the continuous (resp. categorical) features, with ten possible penalization values log-spaced between $10^{-3}$ and $10^{6}$. 
    \item Evaluate the linear model on held out data, using a 10 cross-validation setting (`KFold' cross-validation from scikit-learn). Performance is assessed using \textit{adjusted} accuracy (`balanced\_accuracy\_score' from scikit-learn) for the categorical features, and $R^2$ for the continuous features. Thus, the chance level is zero for both types of features, and the best score is one. 
    \item Report the average decoding performance in Figure \ref{fig:probes} (red bars), and the standard-error of the means across the ten test folds.
\end{itemize}
For comparison, we repeat the exact same procedure with the full GPT-2 activations $X$ (instead of their syntactic component $\overline{X}$), and report the results in Figure \ref{fig:probes} (grey bars). 

\section{Temporal Alignment $g$ between $X$ and $Y$} \label{si:fir}
In Section \ref{sec:mapping}, we map the network's activations $X$ (of length $M$, the number of words) and the brain response $Y$ (of length $N$, the number of fMRI measurements) induced by the same story $w$ (of $M$ words). $M$ is usually greater than $N$. To align the two spaces, we first sum the features between successive fMRI measurements, and then apply a finite impulse response (FIR) model. We denote $g$ this transformation. Specifically, for each fMRI time sample $i \in [1 \dots N]$, $g_i$ combines word features within each acquisition interval as follows:
\begin{align*}\label{equ_align}
    g_i : \bbR^{M \times d} & \rightarrow  \bbR^{5d} \\
    u  & \mapsto 
    \big[ 
    \widetilde{ u_i}, 
    \widetilde{ u_{i-1}}, 
    \dots , 
    \widetilde{ u_{i-4}}
    \big] \\
    \widetilde{ u_i} &= \sum_{\substack{m \in \llbracket 1  \dots  M \rrbracket \\ \mathcal{T}(m) = i}} {u_m} \\
\end{align*}
with
\begin{align*}
\mathcal{T} : \llbracket 1  \dots  M \rrbracket & \rightarrow \llbracket 1  \dots   N \rrbracket \\
m & \mapsto i \quad / \quad | t_{y_i} - t_{x_j} | = \min_{k \in \llbracket1 \dots N\rrbracket} | t_{y_k} - t_{x_m} |
\end{align*}

with $\tilde u$ the summed activations of words between successive fMRI time samples, $u$ the five lags of FIR features, $(t_{x_1}, \dots, t_{x_M})$ the timings of the $M$ words onsets, and $(t_{y_1}, \dots, t_{y_N})$ the timings of the $N$ fMRI measurements.

\section{Brain Parcellation}\label{SI:brain_parcel}
In Figure \ref{fig:rois}, brain scores are averaged across voxels within regions of interest using the Brodmann's areas from the PALS parcellation of freesurfer\footnote{https://surfer.nmr.mgh.harvard.edu/fswiki/PALS\_B12}. To gain in precision, we split the superior temporal gyrus (BA22) into its anterior, middle and posterior parts. In Figure \ref{fig:rois}, we report the top ten areas of the left hemisphere in term of average brain score. Certain areas are renamed for clarity, as specified in the table below:

\begin{tabular}{ll}
\toprule
          Label &                   Corresponding Brodmann's areas \\
\midrule
             A1 &                       BA41 / BA42 \\
       Fusiform &                              BA37 \\
        Angular &                              BA39 \\
           aSTG &                     BA22-anterior \\
           mSTG &                       BA22-middle \\
           pSTG &                    BA22-posterior \\
             M1 &                               BA4 \\
  Supramarginal &                              BA40 \\
       IFG (Op) &                              BA44 \\
      IFG (Tri) &                              BA45 \\
      IFG (Orb) &                              BA47 \\
 Middle-frontal &                              BA46 \\
             V1 &                              BA17 \\
   Fronto-polar &                              BA10 \\
  Temporo-polar &                              BA38 \\
      Precuneus &                               BA7 \\
      Cingulate &  BA23 / BA26 / BA29 / BA30 / BA31 \\
\bottomrule
\end{tabular}


\section{Control for Low-level Linguistic Features}\label{SI:controls}

In Section \ref{sec:res} and Figure \ref{fig:generalisation}, we check that the brain scores are not driven by low-level linguistic features. Thus, we compute the $R$ scores of GPT-2 activations (ninth layer) induced by modified versions of the stimulus:
\begin{itemize}
    \item Random words sampled from the same story. Words are uniformly sampled from the words of the story, tokenized using Spacy \cite{spacy}. Punctuation marks are considered as words. Upper-cases are kept. 
    \item Random sentences from Wikipedia, of the same length as the sentences of the stimulus. We first build a dictionary of (length, list of match-length sentences) pairs out of 10K sentences from Wikipedia ($\approx$ 577K words). Then, for each sentence of the stimulus, a sentence is uniformly sampled from the set of Wikipedia match-length sentences.
    \item The sentences of the stimulus, but with random word order. Words are shuffled \textit{within} each sentence.
\end{itemize}
Then, we extract the corresponding GPT-2 activations and compute the $R$ scores following Section \ref{expe}. $R$ scores are evaluated for each subject and reported in Figure \ref{fig:generalisation}.